\documentclass[letterpaper, 10 pt, conference]{ieeeconf}  % Comment this line out if you need a4paper

\IEEEoverridecommandlockouts                              % This command is only needed if 
\usepackage{dblfloatfix}
\usepackage{multirow}
\usepackage{rotating}
\usepackage{amssymb}
\usepackage{amsmath}
\usepackage[mathscr]{euscript}
\usepackage{tikz}
\def\checkmark{\tikz\fill[scale=0.4](0,.35) -- (.25,0) -- (1,.7) -- (.25,.15) -- cycle;}

\title{\LARGE \bf
CPSeg: Cluster-free Panoptic Segmentation of 3D LiDAR Point Clouds
}
\author{
Enxu Li\textsuperscript{*,2},
Ryan Razani\textsuperscript{*,1},
Yixuan Xu\textsuperscript{2},
and Bingbing Liu\textsuperscript{1}\\
  \textsuperscript{1}Huawei Noah's Ark Lab \ \textsuperscript{2}University of Toronto\\
  \texttt{\{thomas.enxu.li, ryan.razani, richard.xu2 liu.bingbing\}@huawei.com} 
  \thanks{* Indicates equal contribution. \newline This work was done by all authors while at Huawei Noah's Ark Lab.}
}

\begin{document}

\maketitle
\thispagestyle{empty}
\pagestyle{empty}

%%%%%%%%%%%%%%%%%%%%%%%%%%%%%%%%%%%%%%%%%%%%%%%%%%%%%%%%%%%%%%%%%%%%%%%%%%%%%%%%
\begin{abstract}

A fast and accurate panoptic segmentation system for LiDAR point clouds is crucial for autonomous driving vehicles to understand the surrounding objects and scenes. Existing approaches usually rely on proposals or clustering to segment foreground instances. As a result, they struggle to achieve real-time performance. In this paper, we propose a novel real-time end-to-end panoptic segmentation network for LiDAR point clouds, called CPSeg. In particular, CPSeg comprises a shared encoder, a dual-decoder, and a cluster-free instance segmentation head, which is able to dynamically pillarize foreground points according to the learned embedding. Then, it acquires instance labels by finding connected pillars with a pairwise embedding comparison. Thus, the conventional proposal-based or clustering-based instance segmentation is transformed into a binary segmentation problem on the pairwise embedding comparison matrix. To help the network regress instance embedding, a fast and deterministic depth completion algorithm is proposed to calculate the surface normal of each point cloud in real-time. The proposed method is benchmarked on two large-scale autonomous driving datasets: SemanticKITTI and nuScenes. Notably, extensive experimental results show that CPSeg achieves state-of-the-art results among real-time approaches on both datasets.

\end{abstract}

%%%%%%%%%%%%%%%%%%%%%%%%%%%%%%%%%%%%%%%%%%%%%%%%%%%%%%%%%%%%%%%%%%%%%%%%%%%%%%%%
\section{INTRODUCTION}
Recently, panoptic segmentation systems start to draw the attention of the autonomous driving community, since both foreground dynamic objects (i.e. the $\textit{thing}$) and background static scenes (i.e. the $\textit{stuff}$) can be perceived simultaneously.
% Panoptic segmentation systems that predict both semantic tags and instance-level segmentation have recently drawn the attention of the autonomous driving community, for
% because
% both foreground dynamic objects (i.e. the $\textit{thing}$) and background static scenes (i.e. the $\textit{stuff}$) can be perceived and outputted simultaneously. 
However, even though LiDAR is a well-concurred primary perception sensor for autonomous driving for its active sensing nature with high resolution of sensor readings, panoptic segmentation systems using LiDAR point cloud still lack sufficient research compared to image-based approaches.
% Compared with image-based approaches, a panoptic segmentation system using LiDAR point cloud lacks sufficient research yet, despite the fact that LiDAR is a well-concurred primary perception sensor for autonomous driving for its active sensing nature with high resolution of sensor readings. 

% On the leader board of LiDAR panoptic segmentation competition of the well-explored SemanticKITTI dataset \cite{DBLP:conf/iccv/BehleyGMQBSG19}, GP-S3Net \cite{razani2021gps3net} tops the other approaches by the panoptic segmentation metric $\textit{mPQ}$, $\textit{mean of panoptic  quality}$\footnote{By the time of writing this submission}. However, similar to many other state-of-the-arts methods, GP-S3Net is a complicated method that is too difficult to run online, because it is comprised of a complicated semantic segmentation network, AF2-S3Net \cite{cheng2021af2s3net} for foreground object extraction, a heavy-computation clustering module based on density search, and a graph neural network for re-clustering over-segmented clusters into instances.

\begin{figure}[htb!]
    \centering
    \includegraphics[width=0.85\linewidth]{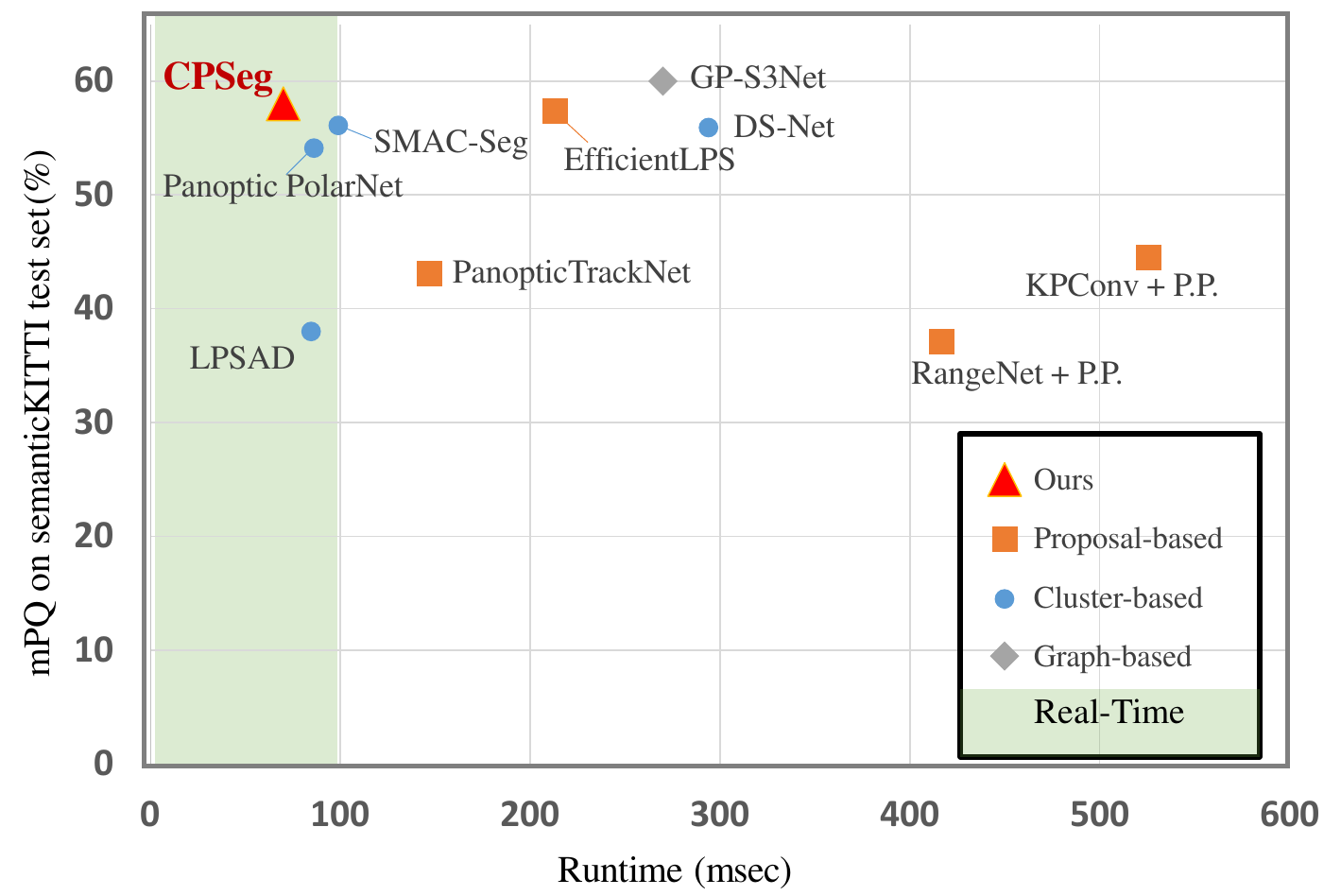}
    \caption{PQ vs runtime on SemanticKITTI \cite{DBLP:conf/iccv/BehleyGMQBSG19} test set. Methods that run in 100 ms or less are considered to be real-time (green area). Our proposed CPSeg outperforms all other available real-time approaches.}
    \label{fig:runtimevsPQ}
    \vspace{-10px}
\end{figure}

GP-S3Net \cite{razani2021gps3net} is the current state of the art in LiDAR panoptic segmentation task. The authors proposed using a 3D sparse convolution-based UNet as a semantic backbone and a combination of HDBSCAN and GCNN to segment instances. However, GP-S3Net is computationally intensive and is thus not a real-time method. Among all published works for LiDAR panoptic segmentation, only few of them \cite{9340837,Zhou2021PanopticPolarNet, li2021smacseg} are capable of operating in real-time (see Figure \ref{fig:runtimevsPQ}). 
% There still exists a performance gap 
A performance gap still exists when comparing real-time methods with the current state of the art.
% A natural question to ask: is it possible to fill this gap and build a panoptic segmentation method with a competitive PQ yet still with a fast running time?
% Therefore, it is important to determine whether it is possible to build a panoptic segmentation system with a competitive PQ but also a fast runtime.
This poses a question: is it possible to build a panoptic segmentation system with a competitive PQ yet still with a fast runtime?

% \vspace{-10px}

Given the runtime constraint for real-time systems, proposal-free methods are favorable as they are more computationally efficient. 
Current proposal-free approaches usually rely on clustering \cite{li2021smacseg,Zhou2021PanopticPolarNet} or graphs \cite{razani2021gps3net} to segment foreground objects. These methods mainly originate from 2D image processing tasks. 
% Current proposal-free approaches in the literature usually rely on clustering or graph to segment foreground objects. These methods are mainly adopted from 2D image processing tasks. 
Designing an effective 3D panoptic segmentation system based on unique characteristics of LiDAR point clouds is still an open research problem.
% However, a LiDAR point cloud has its unique characteristics compare to 2D images where current research in 3D panoptic segmentation (particularly real-time instance segmentation) has not paid much attention to. 
This motivates us to find a more suitable and unique design targeting the panoptic segmentation task in the LiDAR domain. 
% In this work, we take advantage of strong geometric patterns in LiDAR point clouds and present a new proposal-free and cluster-free approach to segment foreground objects with capabilities of running in real-time. 
In this work, we take advantage of geometric patterns in LiDAR point clouds and present a new proposal-free and cluster-free approach to segment foreground objects in real-time. 
In particular, we propose a network to predict object centroid as the embedding of each point and dynamically group points with similar embedding as pillars in the sparse 2D space. Then, objects are formed by building connections of pillars. 

% A natural question to ask: is it possible to build a panoptic segmentation method with a higher mPQ yet still with a fast running time? This work will analyze different components of a panoptic segmentation system and show a new approach for a deploy-able, online LiDAR panoptic segmentation  with a competitive mPQ. 

% There exists other approaches to utilize semantic information for critical, deploy-able perception tasks than panoptic segmentation. For instance, RSN \cite{Sun_2021_CVPR} uses semantic information
% to extract foreground points from LiDAR range images and applies sparse convolutions on the selected, much fewer foreground points to detect object, in a fast and accurate way. However, many LiDAR based panoptic segmentation methods are natural extension from the well-studied LiDAR based semantic segmentation methods in the past few years, due to the fact that semantic segmentation being part of panoptic segmentation.

Our main contributions can be summarized as,
% \begin{enumerate}
%     \item A real-time panoptic segmentation network that is end-to-end (i.e., not relying on deterministic clustering or proposals to segment instances) and achieves state-of-the-art results without extensive post-processing
%     \item A fast surface normal calculation module to aid the process of regressing foreground instance embedding with a novel deterministic depth completion algorithm
%     \item A comprehensive qualitative and quantitative comparison demonstrating the proposed method as opposed to existing methods on both large-scale datasets of SemanticKITTI and nuScenes
%     \item A thorough ablation analysis of how each proposed component contributes to the overall performance
% \end{enumerate}
1) a real-time panoptic segmentation network that is end-to-end (i.e., not relying on deterministic clustering or proposals to segment instances) and achieves state-of-the-art results without extensive post-processing, 
2) a fast surface normal calculation module to aid the process of regressing foreground instance embedding with a novel deterministic depth completion algorithm,
3) a comprehensive qualitative and quantitative comparison between proposed method and existing methods on both large-scale datasets of SemanticKITTI and nuScenes, and
4) a thorough ablation analysis of how each proposed component contributes to the overall performance.

\section{RELATED WORK}

The panoptic segmentation task jointly optimizes semantic and instance segmentation. LiDAR-based semantic segmentation can be categorized into either a projection-based, a voxel-based, or a point-based method depending on the format of data being processed. Projection-based methods project a 3D point cloud into a 2D image plane either in spherical Range-View (RV) \cite{wu2018squeezeseg}, Bird-Eye-View (BEV) \cite{Zhou2021PanopticPolarNet,chen2021polarstream}, or multi-view representations \cite{gerdzhev2021tornadonet}.
Voxel-based methods transform a point cloud into volumetric grids to be processed using 3D convolutions. Processing these 3D grids using 3D convolution is computationally expensive. Therefore, some methods leverage sparse convolutions to alleviate this limitation and to fully exploit sparsity of point clouds \cite{cheng2021af2s3net,zhou2020cylinder3d,tang2020searching}. Point-based methods  \cite{thomas2019kpconv}, however, process the unordered point cloud directly. 
% Despite the high accuracy of the latter approaches, they are inefficient and require large memory consumption. 
Despite having high accuracy, these methods are inefficient and require large memory consumption.
Similar to instance segmentation, panoptic segmentation can be divided into top-down (proposal-based) or bottom-up (proposal-free) methods, as elaborated below. 

\noindent \textbf{Proposal-based panoptic segmentation}
Top-down panoptic segmentation is a two-stage approach. First, foreground object proposals are generated, and subsequently, they are further processed to extract instance information that is fused with background semantic information. Mask R-CNN \cite{He_2017_ICCV} is commonly used for instance segmentation with a light-weight stuff branch segmentation. 
% To resolve the overlapping
% instance predictions by Mask R-CNN several methods are introduced.
To resolve the overlapping
instance predictions by Mask R-CNN and the conflict
between instance and semantic predictions, 
% several methods are introduced.
% Additionally, the conflict
% between instance and semantic predictions are addressed by introducing advance fusion modules. 
% UPSnet \cite{xiong2019upsnet} presents a panoptic head with the
% addition of an unknown class label. 
EfficientPS \cite{sirohi2021efficientlps} proposes to fuse according to their confidence.
Inspired by image-based methods, MOPT \cite{hurtado2020mopt} and EfficientLPS \cite{sirohi2021efficientlps} attach a semantic head to Mask R-CNN to generate panoptic segmentation. \cite{Yin_2021_CVPR} proposes to use center features in regressing 3D bounding boxes and is later adopted to provide instance segmentation results by \cite{fong2021panoptic}.
However, these top-down methods contain
% multiple sequential processes in the pipeline and are usually slow in speed.
multiple slow sequential processes.

\noindent \textbf{Proposal-free panoptic segmentation}
In contrast to proposal-based methods, bottom-up panoptic segmentation predicts semantic segmentation and groups the \textit{thing} points into clusters to achieve instance segmentation. 
% Panoptic-DeepLab \cite{cheng2020panoptic} proposes to predict the instance center locations and group pixels to their closest predicted centers. 
% The pioneering panoptic method in the LiDAR domain, 
LPSAD \cite{9340837} presents a shared encoder with a dual-decoder, followed by a clustering algorithm to segment instances based on the predicted semantic embedding and object centroids. 
% Panoster \cite{gasperini2021panoster} introduces a learnable clustering to assign instance class labels to every point and uses post-processing techniques such as DBSCAN to merge points located close in the 3D space into the same cluster. 
% To the best of our knowledge, Panoster \cite{gasperini2021panoster} is the only cluster-free and proposal-free method in the literature, yet it requires extensive post-processing steps (e.g. using DBSCAN to merge nearby object predictions) to refine the predictions in order to have comparable results with the other state-of-the-art methods.
Panoster \cite{gasperini2021panoster} proposes a learnable clustering module to assign instance class labels to every point. It requires extensive post-processing steps (e.g. using DBSCAN \cite{ester_dbscan} to merge nearby object predictions) to refine the predictions.
% in order to have comparable results with the other state-of-the-art methods. 
DS-Net \cite{hong2020lidar}, however, offers a learnable dynamic shifting module to shift points in 3D space towards the object centroids. GP-S3Net \cite{razani2021gps3net} proposes a graph-based instance segmentation network for LiDAR-based panoptic segmentation. It uses HDBSCAN \cite{campello_hdbscan} to cluster raw point cloud into graph nodes, which can cause confusion in crowded scenes with multiple close-range instances. In contrast, we shift points towards their object centroids using a fully learnable method, improving runtime and performing well even in crowded scenes. Moreover, SMAC-Seg \cite{li2021smacseg} introduces a Sparse Multi-directional Attention Clustering module with a repel loss to better supervise the network separating the instances. However, its usage of large kernels in SMAC introduces additional computation cost. Their method works best with using 0.5m grid size, potentially mixing close-range instances into one grid cell. On the other hand, our dynamic approach of finding connected pillars provides a significant runtime advantage, allowing us to use much finer resolution (e.g. 0.15m grid size) during sub-sampling while retaining real-time performance. 
% allowing us to afford to use much finer grid sizes (e.g. 0.15m) during sub-sampling while retaining real-time performance. The higher resolution allows CPSeg to better detect and segment smaller objects.

\section{PROPOSED METHOD}

\subsection{Problem Formulation}
% \textcolor{red}{Please make sure all the notations are defined and are consistent with the Figures}
Let $(\textbf{P}_{set}, \textbf{L}) = \{ \textbf{p}_{i}, ( c_{i}, o_{i} ) \}^{N}_{i=1} $ be N unordered points of a point cloud where $\textbf{p}_{i} \in \mathbb{R}^{C_0}$ is the input feature for point $i$, tuple $ ( c_{i}, o_{i} ) \in \mathbb{C} \times \mathbb{O}$ is the semantic class label and instance ID label for point $i$. $\mathbb{C}$ is a set of semantic class labels and $\mathbb{O}$ is a set of instance IDs. $\mathbb{C}$ can be further divided into $\mathbb{C}_{thing}$ and $\mathbb{C}_{stuff}$, representing a set of countable foreground thing classes and a set of background stuff classes, respectively. Note that instance label $o_{i}$ is only valid if $c_{i} \in \mathbb{C}_{thing}$. The goal is to learn a function $\mathcal{F}(., \Theta)$, parameterized by $\Theta$, that takes input feature $\textbf{p}_{i}$ and assigns a semantic label for each point and an instance label if it is part of the foreground.

% To solve this problem, we propose CPSeg, an end-to-end network to generate predictions for panoptic segmentation without proposals or clustering algorithms.

\subsection{Network Architecture}
\begin{figure}[htb!]
    \centering
    \includegraphics[width=0.8\linewidth]{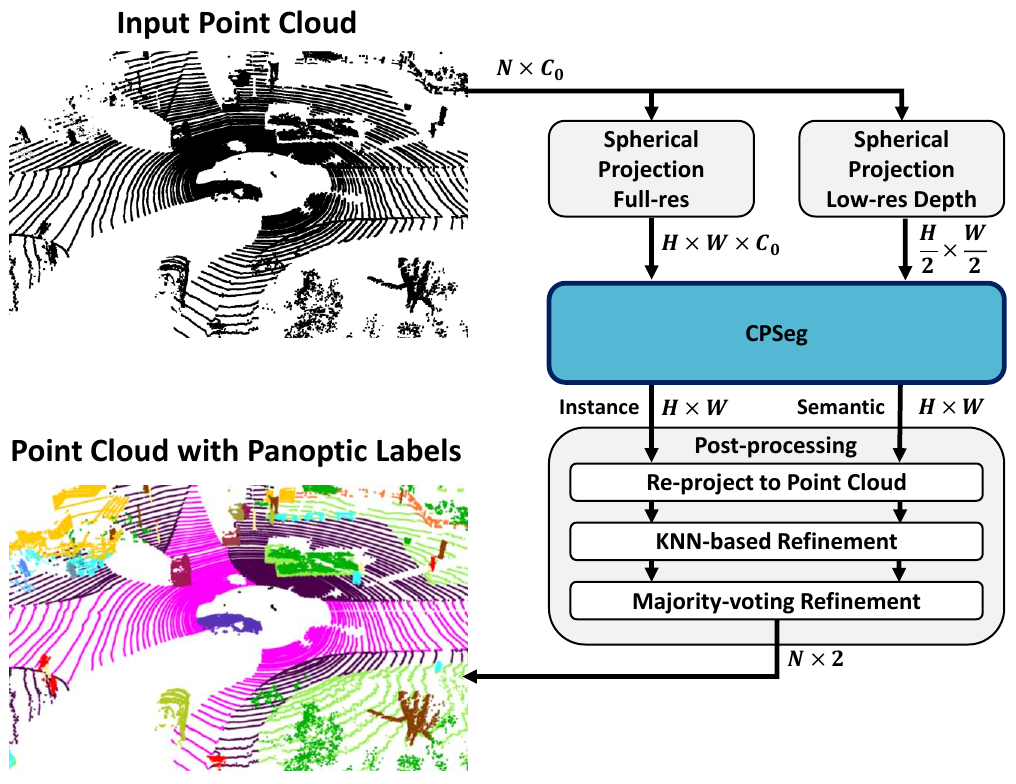}
    \caption{Our method first projects all points into range images to predict instance embedding and semantics. Then, it uses a post-processing to aggregate consistent instances.}
    \label{fig:overview}
    \vspace{-10px}
\end{figure}

The overview of our panoptic segmentation framework is depicted in Figure \ref{fig:overview}. We first transform the LiDAR point cloud $\textbf{P} \in \mathbb{R}^{N \times C_{0}}$ with $C_{0}$ as input features (Cartesian coordinates, remission and depth) into a 2D range image with spatial dimension $H \times W$ using spherical projection similar to \cite{9340837}. At the same time, we build a dense depth map, which will be utilized as a guidance for the depth completion algorithm to extract surface normal features in the following stage. Then, CPSeg takes both inputs and predicts semantic and instance segmentation results in the range view (RV). When re-projecting the results to the 3D point cloud, KNN-based post-processing is utilized to refine the output, as introduced in \cite{milioto2019rangenet++}. Lastly, we fuse the results to obtain panoptic labels and use majority voting to refine the semantic segmentation results where different semantics are predicted in the same instance.

\begin{figure*}
    \centering
    \includegraphics[width=\linewidth]{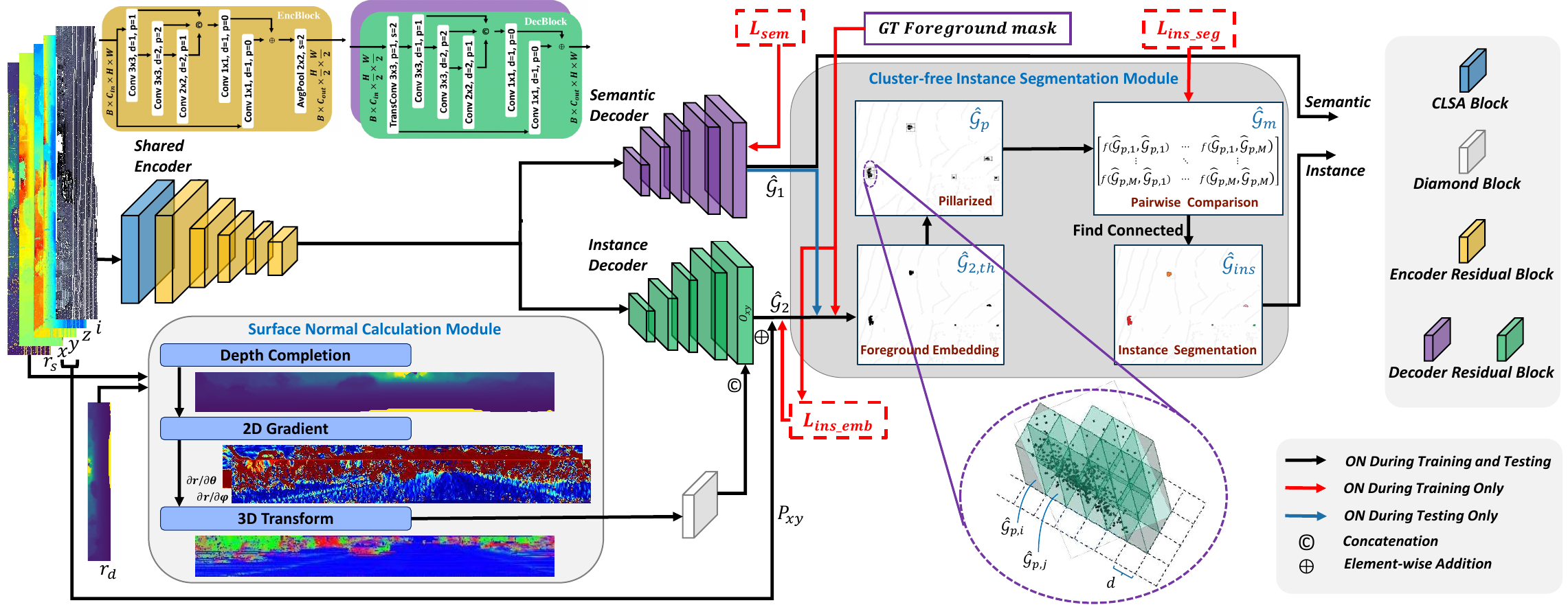}
    % \caption{Illustration of CPSeg. The network takes the projected point cloud as the input and extract features using a shared encoder. The two decoders take the learned features and output semantic segmentation and instance embedding, respectively. The Surface Normal Calculation Module extracts normal feature of the point from the completed depth map which is then fused with the instance features to regress foreground embedding. At the same time, Task-aware Attention Module (TAM) applies channel-wise attention to multi-scale feature maps to enforce the encoder backbone to learn task-aware features for the two decoders. Lastly, the Cluster-free Instance Segmentation Module separates the foreground into pillars and performs a pairwise comparison to find connected nodes which leads to instances. Both semantic and instance predictions are then gathered in RV and sent to post-processing. \textcolor{red}{reduce this caption!}
    \caption{Illustration of CPSeg. The network consists of a dual decoder U-net which processes the input point cloud in RV (with range, Cartesian xyz coordinates, and intensity values) to obtain semantic segmentation and instance embedding. Then, the Cluster-free Instance Segmentation Module separates the foreground into pillars and builds connections which leads to instances. Both semantic and instance predictions are then gathered and sent back to 3D view for post-processing. 
    }
    \label{fig:cpseg}
    \vspace{-10px}
\end{figure*}

Our proposed model is summarized in Figure \ref{fig:cpseg}. It consists of three main components: (A) a dual-decoder U-Net, (B) a surface normal calculation module, which takes the depth maps and computes normal vectors to benefit instance embedding regression, (C) a cluster-free instance segmentation head, which segments the foreground instance embedding into objects.
% \textcolor{red}{A 2D range view representation of a LiDAR point cloud is fed into the CPSeg.or..
% CPSeg takes as input a 2D range view representation of a LiDAR point cloud to be processed by a dual decoder Unet and predicts $\hat{\mathcal{G}}_{1}$, $\hat{\mathcal{G}_{2}}$. }
A 2D RV representation of a LiDAR point cloud is fed into CPSeg.
The output $\hat{\mathcal{G}}_{1}$, $\hat{\mathcal{G}_{2}}$ of the dual-decoder U-Net are the semantic prediction and instance embedding of the projected point cloud, respectively. In particular, the semantic decoder generates $\hat{\mathcal{G}}_{1} \in \mathbb{R}^{H \times W \times C_{class}}$, where $C_{class}$ is the number of semantic classes. With the Cartesian xy coordinates added as a prior, the instance decoder outputs $\hat{\mathcal{G}}_{2} \in \mathbb{R}^{H \times W \times 2} = P_{xy} + O_{xy} $ where $P_{xy}$ is the xy coordinates of the point cloud in RV and $O_{xy}$ is the output from the last block of the instance decoder. Essentially, 
% $\hat{\mathcal{G}}_{2}$ is the instance embedding in 2D space, which could also be interpreted as the predicted 2D location of the object centroids. 
the instance embedding in 2D space, $\hat{\mathcal{G}}_{2}$, is the predicted 2D location of the object centroids.
We then filter using ground truth semantic labels during training or semantic predictions (i.e. $argmax(\hat{\mathcal{G}}_{1})$) during inference to obtain foreground embedding, denoted as $\hat{\mathcal{G}}_{2, th} \in \mathbb{R}^{N \times 2}$ where $N$ is the number of foreground points and 2 refers to the embedding in 2D space.
% Then, a binary mask $\mathcal{M}$ is used to filter foreground points and can be expressed as,
% \vspace{-5px}
% \begin{equation}
%     \mathcal{M} = \mathbb{I} (\mathcal{C} \in \mathbb{C}_{thing}) =
% \begin{cases}
%     1 & \text{if } \mathcal{C} \in \mathbb{C}_{thing}\\
%     0              & \text{otherwise}
% \end{cases}
% \end{equation}
% % \textcolor{red}{Thomas please include the a biref explanation of $I$ :)}
% where $\mathbb{I}$ is the binary conditional function, $\mathcal{C}$ is the ground truth semantic label during training and will be replaced by $argmax(\hat{\mathcal{G}}_{1})$ during test stage. We use $\mathcal{M}$ to obtain corresponding embedding of the foreground \textit{thing} from the instance decoder, denoted as $\hat{\mathcal{G}}_{2, th} \in \mathbb{R}^{N \times 2}$ where $N$ is the number of foreground points and 2 refers to the embedding in 2D space. 
% The foreground embedding is then used by the cluster-free instance segmentation head to segment objects. 

% \subsection{Backbone}
% \textcolor{blue}{we need to better organize the subsections in section 3.. hard to follow. we should talk about shared encoder, and the decoder layers and make a ref to appendix for the detailed architecture..}
\noindent \textbf{Basic architecture} We adopt CLSA module from \cite{li2021smacseg} to extract contextual features. CLSA block learns to recover local geometry in the neighbourhood, which is beneficial for RV-based methods to learn contextual information. The output of the CLSA module is then fed to a shared encoder with five residual blocks, similarly to \cite{cortinhal2020salsanext}, where we obtain multi-scale feature maps, $J_{1}, J_{2}, J_{4}, J_{8}$ (subscript indicates the stride with respect to the full resolution downsampled by \textit{AvgPool} layer at the end of each encoder block). Detailed architecture of encoder and decoder blocks are shown on top left of Figure \ref{fig:cpseg}. Note that each convolution layer is followed by a BatchNorm and a LeakyReLU layer, and that the pooling layer and transposed convolution layer are removed in the last encoder and decoder blocks, respectively.

\subsection{Surface Normal}
% \noindent \textbf{Surface Normal}
Surface normal vectors provide additional geometric cues to regress instance embedding, the shifted 2D location of the object centroid. We use a Diamond Inception Module, adopted from \cite{gerdzhev2021tornadonet}, to extract geometric features from surface normals and directly fuse them with the features in the instance decoder using concatenation operation followed by convolutional layers to obtain instance embedding.
% to regress instance embedding. 
% We provide the final results with and without this module in the Supplementary Materials to further demonstrate its impact to the overall performance.

In this section, we describe a deterministic way of calculating surface normal features of the point cloud with a novel depth completion algorithm. 
% (as depicted in Figure \ref{fig:depthcomplete}). 
The inputs to this module are (A) $\textbf{r}_{s}$, sparse 2D depth map with a scale of $H \times W$, and (B) $\textbf{r}_{d}$, dense 2D depth map with a scale of $\frac{H}{2} \times \frac{W}{2}$. The depth map is obtained by projecting the LiDAR point cloud onto a 2D map with a specified size using discretized indices from spherical transformation, as introduced by \cite{wu2018squeezeseg}.
% \begin{figure}[htb!]
%     \centering
%     \includegraphics[width=\linewidth]{figs/depth_complete.jpg}
%     \caption{Proposed Depth Completion Algorithm. \textcolor{red}{Richard: please change the .jpg with .pdf to be readable}}
%     \label{fig:depthcomplete}
%     % \vspace{-10px}
% \end{figure}
We obtain $\textbf{r}_{row}$, a completed depth map with a weighted row fill using $k$ row neighbours for every entry, as given by, 
% \textcolor{red}{please define all the operations and notations like $\lfloor . \rfloor$, $\lVert . \rVert$. And lets put number for every equation.}
\vspace{-5px}
\begin{equation}
% \[ 
r_{row}[i,j] = \frac{1}{n_{row}[i,j]} \sum_{v=-\lfloor k/2\rfloor}^{\lfloor k/2\rfloor} r_{s}[i,j+v] \cdot o[i,j+v] \cdot w(v)  
% r_{row}[i,j] = \frac{1}{\sum_{v=-\lfloor k/2\rfloor}^{\lfloor k/2\rfloor} o[i,j+v] \cdot w(v)} \\ \sum_{v=-\lfloor k/2\rfloor}^{\lfloor k/2\rfloor} r_{s}[i,j+v] \cdot o[i,j+v] \cdot w(v)  
\label{eq:rowfill}
% \]
\end{equation}
\vspace{-10px}
\begin{equation}
% \[ 
n_{row}[i,j] = \sum_{v=-\lfloor k/2\rfloor}^{\lfloor k/2\rfloor} o[i,j+v] \cdot w(v)  
\label{eq:nfill}
% \]
% \vspace{-10px}
\end{equation}
% \begin{equation}
% % \[ 
% w(v) = ae^{-\frac{v^{2}}{2b^{2}}}  
% % \]
% \end{equation}
where $r_{s}[i,j]$ and $o[i,j]$ are the depth value and binary occupancy at $i^{th}$ row and $j^{th}$ columnm respectively. The operation $\lfloor.\rfloor$ denotes the floor function. The weights $w(v) = ae^{-\frac{v^{2}}{2b^{2}}}$ are sampled from a Gaussian distribution where the center point receives the largest attention and nearby
% the neighboring
points are weighted less as they deviate away from the center. Here, $a$ and $b$ are hyperparameters, which are set to 1 in our method. We then re-use equations \ref{eq:rowfill} and \ref{eq:nfill} to obtain $\textbf{r}_{col}$ using $k$ column neighbours to fill. Next, we bilinear upsample $\textbf{r}_{d}$ to $H \times W$ to obtain a coarse but dense depth map, denoting as $\textbf{r}_{u}$. From $\textbf{r}_{u}$, we calculate $\frac{\partial{\textbf{r}_{u}}}{\partial{\theta}}$ and $\frac{\partial{\textbf{r}_{u}}}{\partial{\phi}}$ using finite difference approximation along horizontal and vertical directions where $\theta$ and $\phi$ are the azimuth and elevation angles for each entry. Local geometry could be interpreted from the two signals. Hence, they serve as the guidance signal to adaptively select a horizontal or vertical fill for each empty entry. For instance, when the magnitude of $\frac{\partial{r_{u}}}{\partial{\phi}}$ is small (i.e. $\lVert\frac{\partial{r_{u}}}{\partial{\phi}}\rVert \rightarrow 0$), it indicates the point is on a pole-like or wall-like object. Therefore, a completion using weighted average of the valid column neighbours is more desired since the change in depth in the vertical direction is relatively small. In summary, each entry in the completed depth map can be expressed as,
\begin{equation}
    r =
\begin{cases}
    r_{s}              & \text{if } o = 1\\
    r_{row} & \text{if } o = 0 \And \lVert\frac{\partial{r_{u}}}{\partial{\theta}}\rVert \le \lVert\frac{\partial{r_{u}}}{\partial{\phi}}\rVert\\
    r_{col}              & \text{if } o = 0 \And \lVert\frac{\partial{r_{u}}}{\partial{\theta}}\rVert > \lVert\frac{\partial{r_{u}}}{\partial{\phi}}\rVert 
\end{cases}
\end{equation}
where index [$i,j$] is omitted for brevity and $o$ denotes occupancy.
From the completed depth map, $\textbf{r}$, we follow \cite{badino2011normal} to calculate the gradients $\frac{\partial{\textbf{r}}}{\partial{\theta}}$ and $\frac{\partial{\textbf{r}}}{\partial{\phi}}$ and transform them into Cartesian frame centered at the LiDAR sensor, obtaining ($\hat{\textbf{n}}_{x}$, $\hat{\textbf{n}}_{y}$, $\hat{\textbf{n}}_{z}$). Note that the purpose of the depth completion algorithm above is to ensure the neighbourhood of valid entry is smooth such that gradients are not influenced by noise.
% \textcolor{red}{We adopt Diamond Inception Module from \cite{gerdzhev2021tornadonet} to extract local geometric features from the calculated surface normal and fuse them with the features in the instance decoder with concatenation followed by convolutions to obtain instance embedding.}

\subsection{Cluster-free Instance Segmentation}

% \textcolor{green}{Read this section more carefully}

Given the 2D embedding of the foreground $\hat{\mathcal{G}}_{2, th}$ 
% $\in \mathbb{R}^{N \times 2}$ 
from the instance decoder, the goal of the cluster-free instance segmentation module is to segment them into instances. First, we dynamically group the foreground points into pillars according to $\hat{\mathcal{G}}_{2, th}$, their location in the 2D embedding space, such that points within grid size $d$ are inside the same pillar (see bottom right of Figure \ref{fig:cpseg}). The embedding of each resulting pillar is the average embedding of the points being grouped together. Pillarized foreground embedding is denoted as $\hat{\mathcal{G}}_{p} \in \mathbb{R}^{M \times 2}$, where M is the number of pillars. Next, we construct a pairwise comparison matrix $\hat{\mathcal{G}}_{m} \in \mathbb{R}^{M \times M}$ to find connected pillars with each entry as,
$\hat{\mathcal{G}}_{m, ij} = f (\hat{\mathcal{G}}_{p, i}, \hat{\mathcal{G}}_{p, j})$, which represents the connectivity probability of pillar $i$ and $j$. A large probability indicates the network is confident that the points in the two pillars belong to the same object. In order for $\hat{\mathcal{G}}_{m}$ to provide meaningful connectivity indications,
we need the function, $f$, to follow several constraints: 
\textbf{1)}. $f (\hat{\mathcal{G}}_{p, i}, \hat{\mathcal{G}}_{p, i}) = 1$, a pillar must be connected to itself with 100\% confidence. 
\textbf{2)}. $f (\hat{\mathcal{G}}_{p, i}, \hat{\mathcal{G}}_{p, j}) = f (\hat{\mathcal{G}}_{p, j}, \hat{\mathcal{G}}_{p, i})$, the connectivity is symmetric; i.e., the network should output the same confidence when comparing pillar $i$ to $j$ and $j$ to $i$. 
\textbf{3)}. $f (\hat{\mathcal{G}}_{p, i}, \hat{\mathcal{G}}_{p, j}) \ge 0$ and $f (\hat{\mathcal{G}}_{p, i}, \hat{\mathcal{G}}_{p, j}) \le 1$, with $f (\hat{\mathcal{G}}_{p, i}, \hat{\mathcal{G}}_{p, j}) \rightarrow 1$ indicating the two pillars belong to the same object.
% \begin{enumerate}
%     \item $f (\hat{\mathcal{G}}_{p, i}, \hat{\mathcal{G}}_{p, i}) = 1$, a pillar must be connected to itself with 100\% confidence.
%     \item $f (\hat{\mathcal{G}}_{p, i}, \hat{\mathcal{G}}_{p, j}) = f (\hat{\mathcal{G}}_{p, j}, \hat{\mathcal{G}}_{p, i})$, the connectivity is symmetric; in other words, the network should output the same confidence when comparing pillar $i$ to $j$ and pillar $j$ to $i$.
%     \item $f (\hat{\mathcal{G}}_{p, i}, \hat{\mathcal{G}}_{p, j}) >= 0$ and $f (\hat{\mathcal{G}}_{p, i}, \hat{\mathcal{G}}_{p, j}) <= 1$, with $f (\hat{\mathcal{G}}_{p, i}, \hat{\mathcal{G}}_{p, j}) \rightarrow 1$ indicating the two pillars belong to the same object.
% \end{enumerate}

We define $f (\hat{\mathcal{G}}_{p, i}, \hat{\mathcal{G}}_{p, j}) = exp(-\alpha \lVert \hat{\mathcal{G}}_{p, i} - \hat{\mathcal{G}}_{p, j} \rVert_{2})$ which satisfies all the constraints listed above. Note that $\alpha$ could be either learned from the pillar features or fixed as a hyperparameter. 
We discuss in detail about the choice of $\alpha$ in the Ablation Studies.
% From experiments, we learned that $\alpha=2.5$ yields the best result.
% We discuss in detail and compare the results in the Supplementary Materials. 
During inference, we use a threshold $T=0.5$ to obtain a binary connectivity matrix, $\hat{\mathcal{G}}_{f}$, formally, $\hat{\mathcal{G}}_{f} = \mathbb{I}(\hat{\mathcal{G}}_{m} > T)$, where $\mathbb{I}$ is the binary conditional function. Note that $\hat{\mathcal{G}}_{f}$ could be interpreted as an adjacency matrix as for the graph structure where each pillar is a node of the graph and a true entry in the matrix represents the two nodes are connected. Then, a simple algorithm \cite{connectedcompoent} is used to find the connected disjoint sets in $\hat{\mathcal{G}}_{f}$ and assign them separate instance IDs. Lastly, we map the pillar instance ID back to the range view using point index matching process. Both semantic and instance segmentation results are now ready to be re-projected back to the point cloud and post-processed.

\subsection{Loss Functions}
\noindent \textbf{Semantic Segmentation Loss}
We follow \cite{gerdzhev2021tornadonet} to supervise the semantic segmentation output, $\hat{\mathcal{G}}_{1}$, with a weighted combination of cross entropy, Lov\'asz softmax, and Total Variation loss, denoted as $L_{sem}$. 
% \begin{equation}
% L_{sem} = \text{WCE}(\hat{\mathcal{G}}_{1}, \mathcal{G}_{1}) + 1.5J(e(\hat{\mathcal{G}}_{1}, \mathcal{G}_{1})) + 7.5\text{TV}(\hat{\mathcal{G}}_{1}, \mathcal{G}_{1})
% \end{equation}
% where $\mathcal{G}_{1}$ is the ground truth (GT) semantic label, $J$ is the Lov\'asz extension of IoU introduced in \cite{berman2018lovasz}, $e(\hat{\mathcal{G}_{1}},\mathcal{G}_{1})$ is the absolute error between the predicted probability and the GT.

\noindent \textbf{Instance Embedding Loss}
We use L2 loss to supervise the learning of instance embedding by taking the difference in predicted instance embedding with the GT, denoted as $L_{ins\_emb}$. Note that the instance embedding here can be interpreted as the mass centroid of an object in 2D BEV.
% \begin{equation}
% L_{ins\_emb} = \sum_{ij} \mathcal{M}_{ij} \cdot \lVert \mathcal{G}_{2,ij} - \hat{\mathcal{G}}_{2,ij} \rVert_{2}
% \end{equation}
% where $ij$ denotes the 2D index on the range image, $\mathcal{M}$ is the GT foreground binary mask to eliminate the background points when calculating loss, $\mathcal{G}_{2}$ is the GT instance embedding, which is the mass centroid of each instance calculated by taking the average of the xy coordinates in the object.

\noindent \textbf{Instance Segmentation Loss}
Essentially, the task here is to supervise binary segmentation on the pairwise matrix $\hat{\mathcal{G}}_{m}$ and optimize the IoUs for positive and negative predictions. Assume points within the same pillar are from the same object, we construct the GT instance label of each pillar by taking the mode label of the points inside, denoted as $\mathcal{G}_{p} \in \mathbb{O}^{M}$ where $\mathbb{O}$ is the set of GT instance labels. The GT binary label for the pairwise comparison matrix, $\mathcal{G}_{m} \in \mathbb{B}^{M \times M}$ is obtained with entries $\mathcal{G}_{m, ij} = \mathbb{I}(\mathcal{G}_{p, i} = \mathcal{G}_{p, j})$.
% \[
% L_{bce} = -\frac{1}{M^{2}} \sum_{i=1}^{M} \sum_{j=1}^{M} \mathcal{G}_{m, ij} log(\hat{\mathcal{G}}_{m, ij}) + (1-\mathcal{G}_{m, ij}) log(1-\hat{\mathcal{G}}_{m, ij})
% \]
\begin{equation}
L_{ins\_seg} = BCE(\hat{\mathcal{G}}_{m},\mathcal{G}_{m}) + J(e(\hat{\mathcal{G}}_{m},\mathcal{G}_{m}))
\end{equation}
where $BCE$ is the binary cross entropy loss, $J$ is the Lov\'asz extension of IoU \cite{berman2018lovasz}, $e(\hat{\mathcal{G}_{m}},\mathcal{G}_{m})$ is the absolute error between the predicted probability and GT. 
% The Lov\'asz loss introduced by \cite{berman2018lovasz} has been shown to be effective in optimizing the IoU metrics. 
% Further experimental results in the Supplementary Materials show adding this loss achieve better overall accuracy.

The total loss that is used to train the network is a weighted combination of the loss terms described above.
\begin{equation}
L_{total} = \beta_{1} L_{sem} + \beta_{2} L_{ins\_emb} + \beta_{3} L_{ins\_seg}
\end{equation}
where $\beta_{1}, \beta_{2}, \beta_{3}$ are the weights for the semantic, instance embedding, and instance segmentation loss terms.

%%%%%%%%%%%%%%%%%%%%%%%%%%%%%%%%%%%%%%%%%%%%%%%%
% SemanticKITTI Test Table Starts HERE
\begin{table*}[htb]
{%\Huge
\centering
% \resizebox{\textwidth}{!}{
\resizebox{1.7\columnwidth}{!}{
\begin{tabular}{l|cccc|ccc|ccc|c|c}
\hline 
Method & PQ  
& PQ\textsuperscript{$\dagger$}
& RQ
& SQ 
& PQ\textsuperscript{Th} 
& RQ\textsuperscript{Th}
& SQ\textsuperscript{Th}  
& PQ\textsuperscript{St}  
& RQ\textsuperscript{St}  
& SQ\textsuperscript{St}  
&  mIoU  
&  FPS  \\
% & \begin{sideways} FPS (Hz) \end{sideways}  \\
\hline
RangeNet++ \cite{milioto2019rangenet++} + PointPillars \cite{Lang_2019_CVPR_pointpillars}
& $37.1$ & $45.9$ & $47.0$ & $75.9$ & $20.2$ & $25.2$ & $75.2$ & $49.3$ & $62.8$ & $76.5$ & $52.4$  & $2.4$\\
PanopticTrackNet \cite{hurtado2020mopt}
& $43.1$ & $50.7$ & $53.9$ & $78.8$ & $28.6$ & $35.5$ & $80.4$ & $53.6$ & $67.3$ & $77.7$ & $52.6$  & $6.8$ \\
KPConv \cite{thomas2019kpconv} + PointPillars \cite{Lang_2019_CVPR_pointpillars}
& $44.5$ & $52.5$ & $54.4$ & $80.0$ & $32.7$ & $38.7$ & $81.5$ & $53.1$ & $65.9$ & $79.0$ & $58.8$ & $1.9$\\
Panoster \cite{gasperini2021panoster}
& $52.7$ & $59.9$ & $64.1$ & $80.7$ & $49.4$ & $58.5$ & $83.3$ & $55.1$ & $68.2$ & $78.8$ & $59.9$  & $-$\\
DS-Net \cite{hong2020lidar}
& $55.9$ & $62.5$ & $66.7$ & $82.3$ & $55.1$ & $62.8$ & $87.2$ & $56.5$ & $69.5$ & $78.7$ & $61.6$  & $3.4^{*}$\\
EfficientLPS \cite{sirohi2021efficientlps}
& $57.4$ & $63.2$ & $68.7$ & $\textbf{83.0}$ & $53.1$ & $60.5$ & $\textbf{87.8}$ & $\textbf{60.5}$ & $\textbf{74.6}$ & $\textbf{79.5}$ & $61.4$  & $4.7$\\
GP-S3Net \cite{razani2021gps3net} 
 & $\textbf{60.0}$ & $\textbf{69.0}$ & $\textbf{72.1}$ & $82.0$ & $\textbf{65.0}$ & $\textbf{74.5}$ & $86.6$ & $56.4$ & $70.4$ & $78.7$ & $\textbf{70.8}$ & $3.7^{*}$
\\
\hline
LPSAD \cite{9340837}
& $38.0$ & $47.0$ & $48.2$ & $76.5$ & $25.6$ & $31.8$ & $76.8$ & $47.1$ & $60.1$ & $76.2$ & $50.9$   & $11.8$ \\
Panoptic-PolarNet \cite{Zhou2021PanopticPolarNet}
 & $54.1$ & $60.7$ & $65.0$ & $81.4$ & $53.3$ & $60.6$ & $\textbf{87.2}$ & $54.8$ & $68.1$ & $77.2$ & $59.5$ & $11.6$
\\
SMAC-Seg \cite{li2021smacseg}
 & $56.1$ & $62.5$ & $67.9$ & $82.0$ & $53.0$ & $61.8$ & $85.6$ & $58.4$ & $72.3$ & $79.3$ & $\textbf{63.3}$  & $10.1$
\\
 \textbf{CPSeg} [\textcolor{blue}{Ours}] 
 & $\textbf{56.9}$ & $\textbf{63.4}$ & $\textbf{68.7}$ & $\textbf{82.3}$ & $\textbf{54.7}$ & $\textbf{63.6}$ & $86.2$ & $\textbf{58.5}$ & $\textbf{72.4}$ & $\textbf{79.4}$ & $62.6$ & $\textbf{14.2}$
\\
\hline
\end{tabular}
}
\caption{Comparison on SemanticKITTI \cite{DBLP:conf/iccv/BehleyGMQBSG19} test dataset. Metrics are provided in [\%] and FPS is in [Hz].(*: source from \cite{li2021smacseg}) }
\label{bigtable} }
\vspace{-10px}
\end{table*} 
% SemanticKITTI Test Table Ends HERE
%%%%%%%%%%%%%%%%%%%%%%%%%%%%%%%%%%%%%%%%%%%%%%%%%

%%%%%%%%%%%%%%%%%%%%%%%%%%%%%%%%%%%%%%%%%%%%%%%%%
% nuScenes Test Table Starts HERE
\begin{table*}[t]
{%\Huge
\centering
% \resizebox{\textwidth}{!}{
% \resizebox{1.6\columnwidth}{!}{
\resizebox{1.7\columnwidth}{!}{
\begin{tabular}{l|cccc|ccc|ccc|c|c}
\hline 
Method & PQ  
& PQ\textsuperscript{$\dagger$}
& RQ
& SQ 
& PQ\textsuperscript{Th} 
& RQ\textsuperscript{Th}
& SQ\textsuperscript{Th}  
& PQ\textsuperscript{St}  
& RQ\textsuperscript{St}  
& SQ\textsuperscript{St}  
&  mIoU  
&  FPS  \\
% & \begin{sideways} FPS (Hz) \end{sideways}  \\
\hline
PanopticTrackNet \cite{hurtado2020mopt}
& $51.6$ & $56.1$ & $63.3$ & $80.4$ & $45.9$ & $56.1$ & $81.4$ & $61.0$ & $75.4$ & $79.0$ & $58.9$  & $-$
\\
EfficientLPS \cite{sirohi2021efficientlps}
& $62.4$ & $66.0$ & $74.1$ & $83.7$ & $57.2$ & $68.2$ & $83.6$ & $71.1$ & $84.0$ & $83.8$ & $66.7$  & $-$
\\
SPVNAS \cite{tang2020searching} + CenterPoint \cite{Yin_2021_CVPR}
& $72.2$ & $76.0$ & $81.2$ & $88.5$ & $71.7$ & $79.4$ & $89.7$ & $73.2$ & $84.2$ & $86.4$ & $76.9$  & $-$
\\

Cylinder3D++ \cite{zhou2020cylinder3d} + CenterPoint \cite{Yin_2021_CVPR}
& $76.5$ & $79.4$ & $85.0$ & $\textbf{89.6}$ & $76.8$ & $84.0$ & $91.1$ & $\textbf{76.0}$ & $\textbf{86.6}$ & $\textbf{87.2}$ & $77.3$  & $-$
\\
(AF)2-S3Net \cite{cheng2021af2s3net} + CenterPoint \cite{Yin_2021_CVPR} 
& $\textbf{76.8}$ & $\textbf{80.6}$ & $\textbf{85.4}$ & $89.5$ & $\textbf{79.8}$ & $\textbf{86.8}$ & $\textbf{91.8}$ & $71.8$ & $83.0$ & $85.7$ & $\textbf{78.8}$  & $-$
\\
\hline
% Dual-Dec UNet w/ BFS [Our Baseline]  
% & $64.1$ & $67.2$ & $74.5$ & $84.8$ & $???$ & $???$ & $???$ & $???$ & $???$ & $???$ & $\textbf{73.7}$  & $7.5$
% \\
% Dual-Dec UNet w/ HDBSCAN [Our Baseline]
% & $66.5$ & $69.5$ & $76.9$ & $85.9$ & $???$ & $???$ & $???$ & $???$ & $???$ & $???$ & $\textbf{73.7}$  & $7.5$
% \\
Panoptic-PolarNet \cite{Zhou2021PanopticPolarNet}
& $63.6$ & $67.1$ & $75.1$ & $84.3$ & $59.0$ & $69.8$ & $84.3$ & $71.3$ & $83.9$ & $84.2$ & $67.0$  & $10.1$
\\
PolarStrean \cite{chen2021polarstream}
& $70.9$ & $74.4$ & $81.7$ & $85.9$ & $70.3$ & $80.3$ & $86.7$ & $71.7$ & $84.2$ & $84.4$ & $69.7$  & $\textbf{22.0}^{\ddag}$
\\
 \textbf{CPSeg} [\textcolor{blue}{Ours}] 
& $\textbf{73.2}$ & $\textbf{76.3}$ & $\textbf{82.7}$ & $\textbf{88.1}$ & $\textbf{72.9}$ & $\textbf{81.3}$ & $\textbf{89.2}$ & $\textbf{74.0}$ & $\textbf{85.0}$ & $\textbf{86.3}$ & $\textbf{73.7}$  & $10.2$

\\
\hline
\end{tabular}
}
\caption{Comparison on nuScenes \cite{fong2021panoptic} test dataset. Metrics are provided in [\%] and FPS is in [Hz]. ($\ddag$: approximated using Figure 5 in \cite{chen2021polarstream})}
\label{bigtable_nu}}
\vspace{-10px}
\end{table*} 
% nuScenes Test Table Ends HERE
%%%%%%%%%%%%%%%%%%%%%%%%%%%%%%%%%%%%%%%%%%%%%%%%%

\section{EXPERIMENTS}

In this section, we describe the experimental settings and evaluate CPSeg on SemanticKITTI \cite{DBLP:conf/iccv/BehleyGMQBSG19} and nuScenes dataset \cite{caesar2020nuscenes} for panoptic segmentation. We compared our results with state-of-the-art approaches. We also provide ablation studies on various components of the network.

\noindent \textbf{Datasets} SemanticKITTI \cite{DBLP:conf/iccv/BehleyGMQBSG19} is the first available dataset on LiDAR-based panoptic segmentation for driving scenes. 
It contains 19,130 training frames, 4,071 validation frames, and 20,351 test frames. 
% We provided ablation analysis and validation results on sequence 08 and test results on sequence 11-21. 
Each point in the dataset is provided with a semantic label of 28 classes, which are mapped to 19 classes for the task of panoptic segmentation. 
Among these 19 classes, 11 belong to \textit{stuff} classes and the rest are considered \textit{things}, where instance IDs are available.

To prove the generalizability of the network, we also benchmarked on the recently announced nuScenes panoptic segmentation dataset \cite{fong2021panoptic}. 
% It is an extension from nuScenes multi-modal dataset \cite{caesar2020nuscenes} 
It contains 700 scenes for training, 150 for validation, and 150 for testing.
10 label classes are considered \textit{things}, and 6 are considered \textit{stuff}.
% Out of the 16 labeled classes for segmentation, 10 countable classes are considered \textit{things}, and 6 other classes are considered \textit{stuff}.
% \cite{caesar2020nuscenes} is another popular large-scale driving-scene dataset, with 700 scenes for training, 150 for validation, and 150 for testing. The data has been primarily collected in four locations in Boston and Singapore, in dense urban environments with many dynamic agents such as different vehicles and pedestrians.
% In the recently released panoptic segmentation dataset \cite{fong2021panoptic} where point-wise instance IDs are made available for the foreground, 10 countable classes are considered \textit{things}, and 6 other classes are considered \textit{stuff}.
% \textcolor{red}{TODO: Change the following: At the time of writing, the authors have not provided point-level panoptic segmentation labels for LiDAR scans. Thus, we generated the labels using the provided 3D bounding box annotations from the \textit{detection} dataset and the semantic labels from the \textit{lidarseg} dataset. In particular, we assign the same instance ID for points within the bounding box with same semantic labels. Out of 16 labeled classes in \textit{lidarseg} datset, 8 human and vehicle classes are considered \textit{things}, and 8 other classes are considered \textit{stuff}. We follow \cite{Zhou2021PanopticPolarNet} to discard instances with fewer than 20 points during evaluation. We train our model on the 700 training scenes and report the results on the validation set of 150 scenes.}

\noindent \textbf{Baselines}
We use a dual-decoder U-Net based on SalsaNext \cite{cortinhal2020salsanext} as the baseline. In particular, the two decoders generate semantic segmentation and instance embedding respectively. Then, a clustering algorithm (e.g. BFS, HDBSCAN) is added after the instance decoder to segment the objects based on the predicted embedding. To be fair in comparison, we add the CLSA Feature Extractor Module in front of the encoder to match our network design. Moreover, we implement LPSAD based on \cite{9340837} as an additional baseline. Quantitative and qualitative results are compared against the proposed methods on the SemanticKITTI and nuScenes validation set.

\noindent \textbf{Evaluation Metric} We follow \cite{kirillov2019panoptic} to use the mean Panoptic Quality (PQ) as our main metric to evaluate and compare the results with others. In addition, we also report Recognition Quality (RQ), and Segmentation Quality (SQ). They are calculated separately on \textit{stuff} and \textit{thing} classes, providing PQ\textsuperscript{St}, SQ\textsuperscript{St}, RQ\textsuperscript{St} and PQ\textsuperscript{Th}, SQ\textsuperscript{Th}, RQ\textsuperscript{Th}.

%%%%%%%%%%%%%%%%%%%%%%%%%%%%%%%%%%%%%%%%%%%%%%%%%
%%% Qualitative result
%%%%%%%%%%%%%%%%%%%%%%%%%%%%%%%%%%%%%%%%%%%%%%%%%

\begin{figure}[t]
    \centering
    \includegraphics[width=\linewidth]{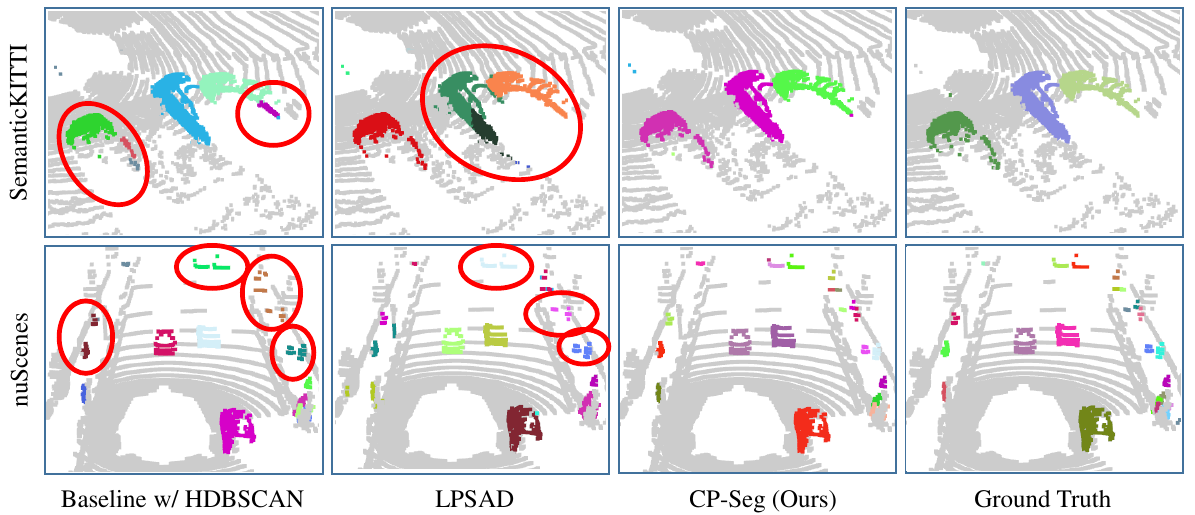}
    % \caption{Qualitative comparison of CPSeg with other methods on both SemanticKITTI and nuSenes validation set.}
    \caption{Qualitative comparison of CPSeg with other methods.}
    \label{fig:qualitative}
    \vspace{-10px}
\end{figure}
%%%%%%%%%%%%%%%%%%%%%%%%%%%%%%%%%%%%%%%%%%%%%%%%%
% \vspace{-10px}

\subsection{Experimental Setup}
For both datasets, we trained CPSeg end-to-end for 150 epochs using SGD optimizer and exponential-decay learning rate scheduler with initial learning rate starting at 0.01 and a decay rate of 0.99 every epoch. A weight decay of $10^{-4}$ was used. The model was trained on 4 NVIDIA V100 GPUs with a batch size of 4 per GPU. The weights for the losses were set to $\beta_{1}=1.0$, $\beta_{2}=0.1$, $\beta_{3}=0.2$. We used a range image with resolution of ($H=64$, $W=2048$). 
% The pillar grid size, $d$ of the final models was set to $0.15$ with pillar pairwise matrix threshold, $T$ to be $0.5$. The mapping parameter, $\alpha$ was set to be $2.5$ for the final model.

\subsection{Quantitative Evaluation}
In Table \ref{bigtable} and Table \ref{bigtable_nu}, we compile the results of CPSeg compared to other models, separating the models into two groups based on their inference speed. For evaluations on SemanticKITTI test dataset (Table \ref{bigtable}), only the models in row 8-11 are known to have real-time performance, with FPS's above 10Hz. With a PQ of $56.9\%$ and an FPS of 14.2Hz,  CPSeg achieves performances that match state-of-the-art models. More importantly, it establishes a new benchmark in PQ for real-time models, surpassing the next best real-time model, SMAC-Seg, by $0.8\%$. Specifically, with a $1.8\%$ increase in {RQ\textsuperscript{Th}} and 3.1Hz improvement in FPS over SMAC-Seg, we demonstrate that CPSeg is better in recognizing foreground objects while using less computation. These improvements can be mainly attributed to the use of cluster-free instance segmentation module and the incorporation of surface normal to aid embedding regression. 
% as a helpful part of the instance embedding.

% This higher mIoU and SQ can be mainly attributed to the CLSA module, which incorporates 3D geometry features that are otherwise lost after spherical projection.

% For results on nuScenes validation dataset (Table \ref{bigtable_nu}), since the methods for creating the instance labels are not standardized across publications, we separated the models into three groups for better comparison. The models in rows 1-8 are previously published models, grouped by inference speed similar to Table \ref{bigtable}. The baseline and proposed models listed in row 9-12 use results from our experiments. Using $64 \times 2048$ range images as input, CPSeg HR (row 12) obtains the highest PQ out of all models, outperforming the baseline models Dual Decoder (BFS) and Dual Decoder (HDBSCAN) by $18.5\%$ and $7.8\%$, respectively. 
On nuScenes test dataset (Table \ref{bigtable_nu}), CPSeg again achieves competitive segmentation ability. Notably, it obtains the highest PQ for models with real-time performances, outperforming the PolarStream and Panoptic-PolarNet by $2.3\%$ and $9.6\%$. Although Cylinder3D++ and (AF)2-S3Net, when combined with CenterPoint, achieve better PQ, their feature extraction based on 3D voxels is much more time-consuming compared to the 2D feature extraction in CPSeg.
% We also avoided comparison of results on nuScenes validation dataset, since the methods for creating the instance labels were not standardized across publications. 
% Thus, we compare the baseline and proposed models used in our experiments. Despite achieving the same semantic segmentation performance in mIoU, CPSeg (row 3) proves to be superior in differentiating instances. It obtains the highest PQ out of all models, outperforming the baseline models Dual Decoder (BFS) and Dual Decoder (HDBSCAN) by $9.1\%$ and $6.7\%$, respectively. Note that CPSeg operates slightly slower on nuScenes since the setting is more crowded compared to SemanticKITTI dataset with more instances in each scene, yet it is still real-time. 
% By reducing the resolution from $64 \times 2048$ to $32 \times 1024$, CPSeg (row 11) again achieves a competitive real-time performance with only a small trade-off in PQ.

\label{sec:quanti}

\subsection{Qualitative Evaluation}
\label{sec:quali}

% The panoptic segmentation performance 
The performance
of CPSeg can also be seen in Figure \ref{fig:qualitative}, where we compare its inference results to LPSAD, our implementation based on \cite{9340837}, and baseline models. For a closedup view of a scene from SemanticKITTI dataset (row 1) where three cars are lined up closely, only CPSeg segments the instance points without errors. LPSAD identifies the car in the middle as two separate instances, whereas the baseline model produces even worse over-segmentation errors.

In a complex scene from nuScenes (row 2), with variations in instance classes and few sparse points describing each instance, correctly recognizing and distinguishing each instance is proved to be more difficult. For areas where pedestrians walk closely or where cars are positioned further away, the baseline model using HDBSCAN and LPSAD are prone to making under-segmentation errors. In such a complex scene, only CPSeg is able to segment accurately.

\subsection{Ablation Studies}
In this section, we present an extensive ablation analysis on proposed components in CPSeg. Note that all results are compared on SemanticKITTI validation set (Seq 08). First, we investigate the individual contribution of each component in the network, as shown in Table \ref{tab:Ablation}. The cluster-free instance segmentation module is the key component, introducing $9.6\%$ increase on the PQ (compare to the baseline with BFS) while eliminating the computation of clustering.
% at the same time
Moreover, extracting surface normal brings another jump in PQ since the network receives guidance on regressing the embedding for each foreground object. Lastly, the model achieves the best result by incorporating binary Lov\'asz loss in supervising the segmentation on the pairwise matrix.

\begin{table}[h]
\begin{center}
% \scalebox{0.6}
\scalebox{0.7}
{
\begin{tabular}{ c|ccc c  }
\hline %\hline %\\
\multicolumn{1}{c|}{\textbf{Architecture}} &
\multicolumn{1}{c}{\textbf{\begin{turn}{45} Cluster-free \end{turn} }} &
% \multicolumn{1}{c}{\textbf{\begin{turn}{45} TAM \end{turn} }} &
\multicolumn{1}{c}{\textbf{\begin{turn}{45} 3D Normal \end{turn} }} &
\multicolumn{1}{c}{\textbf{\begin{turn}{45} Lov\'asz \end{turn} }} &
\multicolumn{1}{|c}{\textbf{mPQ}} \\

 \hline \hline 
\multirow{1}{*}{Baseline w/ BFS}
&       &       &   \multicolumn{1}{c|}{ } & 44.9 \rule{0pt}{3ex}\\%\hline
\multirow{1}{*}{Baseline w/ HDBSCAN}
&      &       &   \multicolumn{1}{c|}{ } & 52.7 \rule{0pt}{3ex}\\\hline
\multirow{6}{*}{Proposed}
& \checkmark &           & \multicolumn{1}{c|}{ }  & $54.5$  \rule{0pt}{3ex}\\

% & \checkmark & \checkmark &\checkmark  &   & \multicolumn{1}{c|}{ }  & dd  \rule{0pt}{3ex}\\ %\cline{2-7}  %\hline

& \checkmark  &\checkmark   &       \multicolumn{1}{c|}{ }  & $55.6$  \rule{0pt}{3ex}\\ %\cline{2-7}  %\hline

& \checkmark  &   &       \multicolumn{1}{c|}{\checkmark}  & $55.3$  \rule{0pt}{3ex}\\ %\cline{2-7}  %\hline

& \checkmark &\checkmark      &    \multicolumn{1}{c|}{\checkmark}  & $\textbf{56.2}$  \rule{0pt}{3ex}\\ \hline

\end{tabular}}
\end{center}
% \vspace{-10px}
\caption{Ablation study of the proposed model with individual components vs baseline. Metrics are provided in [\%].}
\label{tab:Ablation}
\vspace{-10px}
\end{table}
\vspace{-10px}
We experiment with changing $\alpha$, the parameter used to map the pillar embedding to the connectivity probability. We set the threshold $T$ to be 0.5, and pillar grid size $d$ to be 0.15 for the experiments on $\alpha$. In the first setting, $\alpha$ is learned from the corresponding pillar feature from the instance decoder. In particular, $\alpha = MLP(F_{ins, i} \mathbin\Vert F_{ins, j})$, where $F_{ins, i}$ and $F_{ins, j}$ are the corresponding features of pillar $i$ and $j$ from the instance decoder, and $\mathbin\Vert$ denotes the concatenation operation. In the second setting, we set $\alpha$ to be various fixed values. From the results in Table \ref{tab:alpha}, constant $\alpha=2.5$ yields the best results. A fixed value works relatively better than learning from the feature; for panoptic segmentation tasks on outdoor autonomous driving dataset, the difference in the regressed 2D embedding is enough to determine the connectivity of the pillars. However, we think that a learned $\alpha$ could potentially work better if the scene is dense and crowded (indoor scenes) such that the network requires more information in making connections. Also note that $\alpha=2.5 \approx -ln(T) / (2 \times d)$. We draw conclusions that $\alpha$ can be regarded as a function of the threshold, $T$, and pillar grid size, $d$. Hence, we choose $\alpha$ to be $\frac{-ln(T)}{2d}$ for the rest of the experiments. This design choice ensures that adjacent pillars are considered to be connected.

\begin{table}[htb!]
\begin{center}
\scalebox{0.8}
{
\begin{tabular}{ c|c|cccc  }
\hline %\hline %\\
\multicolumn{1}{c|}{} &
\multicolumn{1}{c|}{$\alpha$} &
\multicolumn{1}{c}{PQ} &
\multicolumn{1}{c}{PQ\textsuperscript{Th}} &
\multicolumn{1}{c}{RQ\textsuperscript{Th}} &
\multicolumn{1}{c}{SQ\textsuperscript{Th}} \\

 \hline \hline 
 \multirow{4}{*}{Fixed}
&$0.1$ & $44.7$ & $31.5$ & $38.7$      &    $71.9$   \rule{0pt}{3ex}\\ \cline{2-6}%\hline
%  \multirow{1}{*}{0.05}
&$1.0$ & $56.0$ & $58.4$ & $66.2$      &    $\textbf{76.7}$   \rule{0pt}{3ex}\\ \cline{2-6}%\hline
 
% \multirow{1}{*}{0.15}
&$2.5$ & $\textbf{56.2}$ & $\textbf{58.7}$ & $\textbf{66.6}$      &    $76.5$   \rule{0pt}{3ex}\\ \cline{2-6}%\hline

% \multirow{1}{*}{0.3}
&$5.0$ & $48.7$ & $41.0$ & $51.2$      &    $68.0$  \rule{0pt}{3ex}\\ \hline

\multirow{1}{*}{Learned}
&$-$ & $55.3$ & $56.2$ & $64.9$      &    $75.5$   \rule{0pt}{3ex}\\ \hline

% \multirow{1}{*}{0.5}

\end{tabular}
}
\end{center}
% \vspace{-10px}
\caption{Ablation study of using different $\alpha$. Metrics provided in [\%].}
\label{tab:alpha}
\vspace{-10px}
\end{table}
\vspace{-10px}
One may concern about the complexity of the model as it grows quadratically with $M$, the number of pillars. Note that we provide the average number of pillars resulted from using different grid sizes in Table \ref{tab:pillarsize}. Typically, a SemanticKITTI LiDAR scan contains an average number of 12 instances and 6.8k number of foreground points. We find that $M$ is proportional to the number of instances in the scan but significantly less than the number of points. As the point embedding is learned to shift together in the network, dynamically grouping the foreground points together using pillars according to their embedding significantly reduces the computation the network needs to carry. 
% In fact, the main contribution that results in CPSeg's real-time performance comes from the cluster-free instance segmentation module. 
The baselines with BFS clustering (row 1 in Table \ref{tab:Ablation}) and HDBSCAN (row 2 in Table \ref{tab:Ablation}) run in 8.6Hz and 4.8Hz respectively. In contrast, CPSeg runs in 14.2Hz. By dynamically grouping LiDAR points with similar embedding, CPSeg only processes on average 141 pillars instead of thousands of LiDAR points.

\begin{table}[htb!]
\begin{center}
\scalebox{0.8}
{
\begin{tabular}{ c|c|cccc|c  }
\hline %\hline %\\
\multicolumn{1}{l|}{Grid Size (m)} &
\multicolumn{1}{l|}{$M$} &
\multicolumn{1}{c}{PQ} &
\multicolumn{1}{c}{PQ\textsuperscript{Th}} &
\multicolumn{1}{c}{RQ\textsuperscript{Th}} &
\multicolumn{1}{c}{SQ\textsuperscript{Th}}  &
\multicolumn{1}{|c}{Runtime (ms)} \\

 \hline \hline 
%  \multirow{1}{*}{0.05}
$0.05$ & $430$ & $55.1$ & $56.7$ & $65.6$      &    $75.0$  &    $78$ \rule{0pt}{3ex}\\ \hline
 
% \multirow{1}{*}{0.15}
$0.15$ &$141$ & $\textbf{56.2}$ & $\textbf{58.7}$ & $\textbf{66.6}$      &    $76.5$  &    $70$ \rule{0pt}{3ex}\\ \hline

% \multirow{1}{*}{0.3}
$0.30$ &$103$ & $56.0$ & $57.6$ & $65.7$      &    $76.2$ &    $68$  \rule{0pt}{3ex}\\ \hline

% \multirow{1}{*}{0.5}
$0.50$ &$91$ & $55.8$ & $57.6$ & $65.4$      &    $\textbf{76.6}$ &    $\textbf{67}$  \rule{0pt}{3ex}\\ \hline

\end{tabular}
}
\end{center}
% \vspace{-10px}
\caption{Ablation study of using different grid sizes in pillarizing foreground points by embedding. Metrics provided in [\%].}
\label{tab:pillarsize}
\vspace{-10px}
\end{table}
\vspace{-10px}
% Generally, using a finer grid size increases the resolution of the BEV pillar map, allowing for a more detailed instance segmentation prediction and thus a higher PQ. However, we find the performance drops when the grid size is below 0.15m. 
% % the instance embedding used for pillarization is not perfect.
% % Some difficult points could lead to inaccurate instance embedding predictions.
% During inference, some background points could be included due to semantic errors, and their instance embedding is not meaningful. 
% These errors are less pronounced when the grid size is large and more points are incorporated to compute the average embedding, but more pronounced when the grid size shrinks. For this reason, PQ is lower when the grid size changes from 0.15m to 0.05m. 
\section{CONCLUSION}
In this work, we propose a novel real-time proposal-free and cluster-free panoptic segmentation network for 3D point cloud, called CPSeg.
Our method builds upon an efficient semantic segmentation network and addresses the instance segmentation by incorporating a unique cluster-free instance head where the foreground point cloud is dynamically pillarized in the sparse space according to the learned embedding and object instances are formed by building connection of pillars. 
% Moreover, a novel task-aware attention module is designed to enforce two decoders to learn task-specific features. 
CPSeg outperforms existing real-time LiDAR-based panoptic segmentation methods on both SemanticKITTI and nuScenes datasets. 
% CPSeg outperforms existing LiDAR-based panoptic segmentation methods in terms of inference time and is on-par with state-of-the art on both datasets of SemanticKITTI and nucSenes. 
% \textcolor{red}{Instead, We can simply say: CPSeg outperforms existing real-time LiDAR-based panoptic segmentation methods on both datasets of SemanticKITTI and nucSenes. or CPseg achieves state-of-the-art performance among existing real-time methods on both...}
The thorough analysis illustrates the robustness and effectiveness of the proposed method, which could inspire the field and push the panoptic segmentation research towards a proposal-free and cluster-free direction.

%%%%%%%%%%%%%%%%%%%%%%%%%%%%%%%%%%%%%%%%%%%%%%%%%%%%%%%%%%%%%%%%%%%%%%%%%%%%%%%%
% \begin{thebibliography}{99}
% \bibliography{bib}
% \end{thebibliography}

\bibliographystyle{IEEEtran}
\bibliography{bib}

\end{document}